\documentclass[conference]{IEEEtran}
\IEEEoverridecommandlockouts
\usepackage{cite}
\usepackage{amsmath,amssymb,amsfonts}
\usepackage{algorithmic}
\usepackage{graphicx}
\usepackage{textcomp}
\usepackage{xcolor}
\usepackage{hyperref}
\usepackage{booktabs}
\usepackage{multirow}
\usepackage{cleveref}

\def\BibTeX{{\rm B\kern-.05em{\sc i\kern-.025em b}\kern-.08em
    T\kern-.1667em\lower.7ex\hbox{E}\kern-.125emX}}

\begin{document}

\title{Zero-shot System for Automatic Body Region Detection for Volumetric CT and MR Images}

\author{\IEEEauthorblockN{Farnaz Khun Jush*}
\IEEEauthorblockA{\textit{Radiology} \\
\textit{Bayer AG}\\
Berlin, Germany \\
farnaz.khunjush@bayer.com}
\and
\IEEEauthorblockN{Grit Werner}
\IEEEauthorblockA{\textit{Radiology} \\
\textit{Bayer AG}\\
Berlin, Germany \\
grit.werner@bayer.com}
\and
\IEEEauthorblockN{Mark Klemens}
\IEEEauthorblockA{\textit{Radiology} \\
\textit{Bayer AG}\\
Berlin, Germany \\
mark.klemens@bayer.com}
\and
\IEEEauthorblockN{Matthias Lenga}
\IEEEauthorblockA{\textit{Radiology} \\
\textit{Bayer AG}\\
Berlin, Germany \\
matthias.lenga@bayer.com}
\thanks{*Farnaz Khun Jush is the corresponding author.}
}

\maketitle

\begin{abstract}
Reliable identification of anatomical body regions is a prerequisite for many automated medical imaging workflows, yet existing solutions remain heavily dependent on unreliable DICOM metadata. Current solutions mainly use supervised learning, which limits their applicability in many real-world scenarios. In this work, we investigate whether body region detection in volumetric CT and MR images can be achieved in a fully zero-shot manner by using knowledge embedded in large pre-trained foundation models. We propose and systematically evaluate three training-free pipelines: (1) a segmentation-driven rule-based system leveraging pre-trained multi-organ segmentation models, (2) a Multimodal Large Language Model (MLLM) guided by radiologist-defined rules, and (3) a segmentation-aware MLLM that combines visual input with explicit anatomical evidence. 
All methods are evaluated on 887 heterogeneous CT and MR scans with manually verified anatomical region labels. The segmentation-driven rule-based approach achieves the strongest and most consistent performance, with weighted F1-scores of 0.947 (CT) and 0.914 (MR), demonstrating robustness across modalities and atypical scan coverage. The MLLM performs competitively in visually distinctive regions, while the segmentation-aware MLLM reveals fundamental limitations.

\indent \textit{Clinical relevance}—This study demonstrates that accurate and metadata-independent body region detection is feasible without any task-specific training. The proposed framework requires no training data, is adaptable to evolving anatomical definitions, and is suitable for clinical and research workflows.
\end{abstract}


\section{INTRODUCTION}

Recent advancements in computer-aided diagnosis (CAD) and medical image analysis have revolutionized clinical workflows by automating tasks such as organ segmentation \cite{cardenas2019advances,fang2020multi,pan2023abdomen}, and anatomical landmark identification \cite{zhou2021learn, zhu2021you}. At the core of many applications lies the accurate detection of body regions in medical scans, ensuring that downstream processes such as lesion detection, treatment planning, image-guided applications, quality controls, etc., are anchored in anatomically relevant areas \cite{golzan2025automatic}. 
However, traditional methods on DICOM metadata are inherently unreliable. Metadata errors stem from many sources, e.g., manual entry mistakes, inconsistent institutional naming conventions, and vendor-specific protocol variations \cite{gueld2002quality,jush2025content}. Such inaccuracies risk misdirecting downstream tasks and potentially delaying or failing automated methods, for instance, a mislabeled abdominal scan might trigger a liver segmentation model to analyze unrelated anatomy.

Pixel-based methods address these limitations by directly analyzing anatomical content within images, bypassing error-prone metadata. These approaches leverage intrinsic visual features such as tissue density gradients, organ morphology, and spatial relationships to classify body regions with high precision \cite{raffy2023deep, ouyang2022deep,yan2022sam}. Pixel-based models are adaptable to diverse imaging protocols and modalities (e.g., contrast-enhanced vs. non-contrast scans, MRI vs. CT) \cite{raffy2023deep, yan2022sam}. 
Automated body region detection can reduce reliance on radiologists' manual curation, freeing clinicians to focus on complex diagnostic tasks rather than time-consuming data preprocessing. 
The development of flexible, robust body region detection models is thus critical. Such systems must generalize across anatomical variability while maintaining computational efficiency for real-world deployment \cite{hrzic2025simple, raffy2023deep}.

\subsection{Prior Work}

Despite the critical role of body region detection in enhancing clinical workflows, this area remains understudied compared to well-established tasks like organ segmentation or lesion classification \cite{golzan2025automatic}. 
Recent studies addressed the problem of body region detection by employing supervised deep learning methods, such as training convolutional neural networks (CNNs) to classify anatomical regions \cite{raffy2023deep, hrzic2025simple, golzan2025automatic}. Their development relies on large labeled datasets. 
Leveraging pre-trained segmentation models such as TotalSegmentator \cite{wasserthal2023totalsegmentator, d2024totalsegmentator}, which robustly segment multiple organs and modalities (e.g. CT, MRI) across diverse anatomical regions can be a way to metigiate the data requirement challenge. These models, trained on vast datasets, provide a foundation for transfer learning or zero-shot inference, where knowledge from one task is adapted to another without extensive retraining. For instance, features learned from multi-organ segmentation could be repurposed to infer body regions, reducing dependency on region-specific labeled data \cite{jush2025content, khun2025content}.

Parallel advancements in Multimodal Large Language Models (MLLMs) further expand possibilities. MLLMs have demonstrated a remarkable ability to interpret medical images by correlating visual patterns with textual context, suggesting potential for metadata-agnostic body region detection \cite{urooj2025large}.

\subsection{Problem Statement and Contribution}

Reliable body region detection remains a challenge in medical imaging, particularly due to the scarcity of high-quality region-level annotations and the frequent unreliability of DICOM metadata.  
A key open question is whether body region detection can be performed without any dedicated training, by reusing knowledge already embedded in large pre-trained models.
Thus, here we investigate whether body region detection in CT and MRI scans can be achieved in a zero-shot manner. The core challenge is to design a system that is flexible, explainable, and adaptable, capable of inferring visible body regions solely from anatomical content while remaining robust to protocol variations and capable of generalizing across datasets.
The contribution of this work is as follows:

\begin{itemize} 
\item A novel zero-shot, segmentation-driven framework with rule-based reasoning for flexible and interpretable body region detection. 
\item A zero-shot approach leveraging MLLMs with rule-based prompting. 
\item A hybrid zero-shot model integrating segmentation-derived anatomical evidence into MLLMs to combine visual and explicit anatomical information. 
\item A systematic comparison of zero-shot paradigms for volumetric body region detection across heterogeneous CT and MRI datasets, including a comparison to a state-of-the-art supervised baseline \cite{golzan2025automatic}. 
\end{itemize}

\section{Methods}

This study investigates three complementary zero-shot strategies for anatomical body region detection in volumetric CT and MRI scans. 
All methods operate in a fully zero-shot setting, without any body-region-specific training, fine-tuning, or use of DICOM metadata. 
Instead, they exploit anatomical knowledge embedded in large pre-trained foundation models: a multi-organ segmentation model \cite{wasserthal2023totalsegmentator} and a general-purpose MLLM (here Gemini 2.5 \cite{comanici2025gemini}). The three evaluated pipelines are: (i) a segmentation-driven rule-based system, (ii) an MLLM system using visual reasoning, and (iii) a hybrid segmentation-aware MLLM system.
To ensure consistency and reproducibility, we first define the anatomical regions and boundaries. We then provide a detailed explanation of each zero-shot pipeline.

\subsection{Anatomical Region Definition}

\begin{figure}[h]
    \centering
    \includegraphics[trim=2.5cm 1cm 2.5cm 0cm, clip, width=\columnwidth]{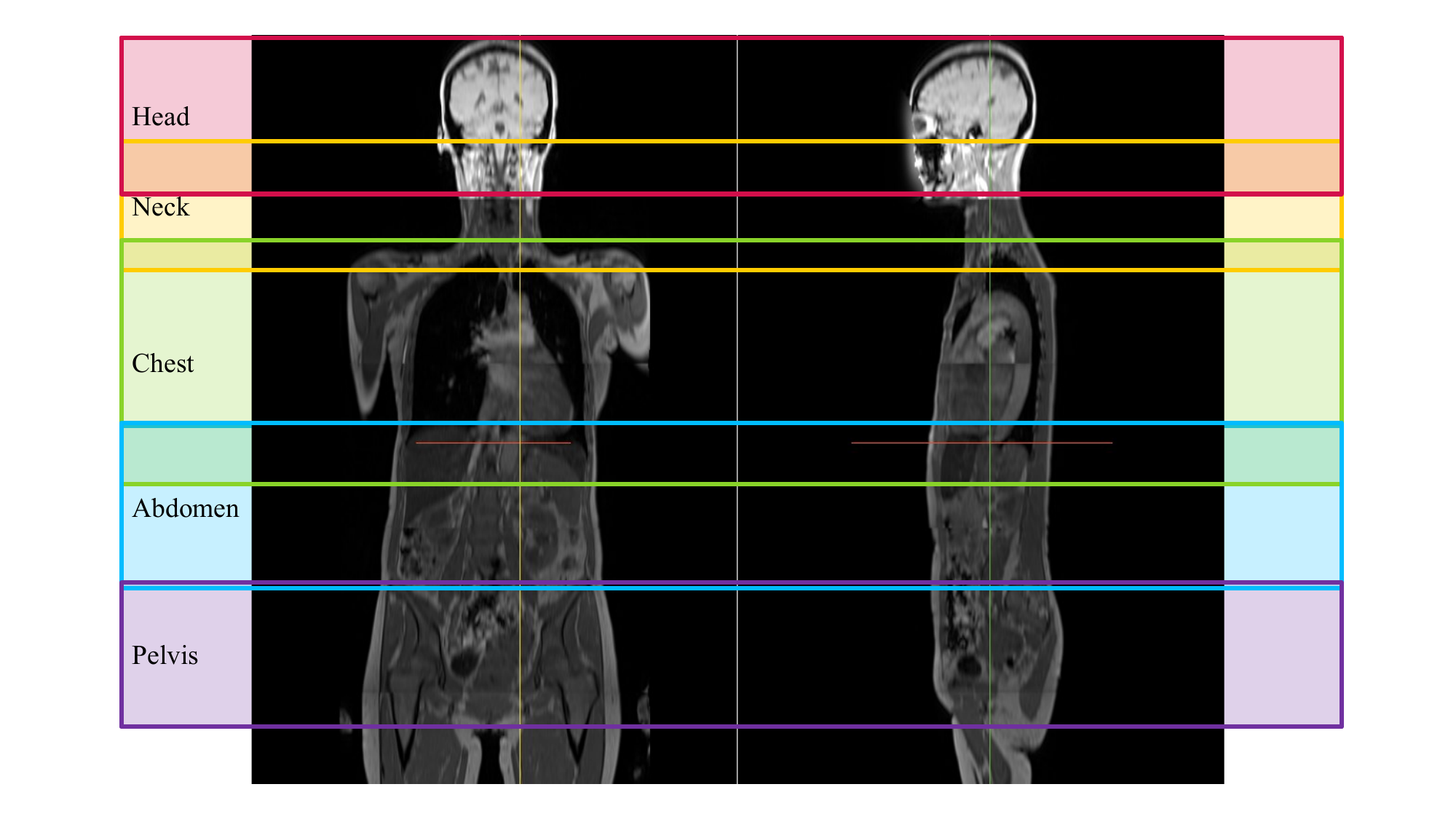}
    \caption{Example anatomical regions delineated on a whole body MR Image (sagittal and coronal view).}
    \label{fig: CT regions}
\end{figure}

\begin{figure}[h]
    \centering
    \includegraphics[trim=2.5cm 0.5cm 2.5cm 0cm, clip, width=\columnwidth]{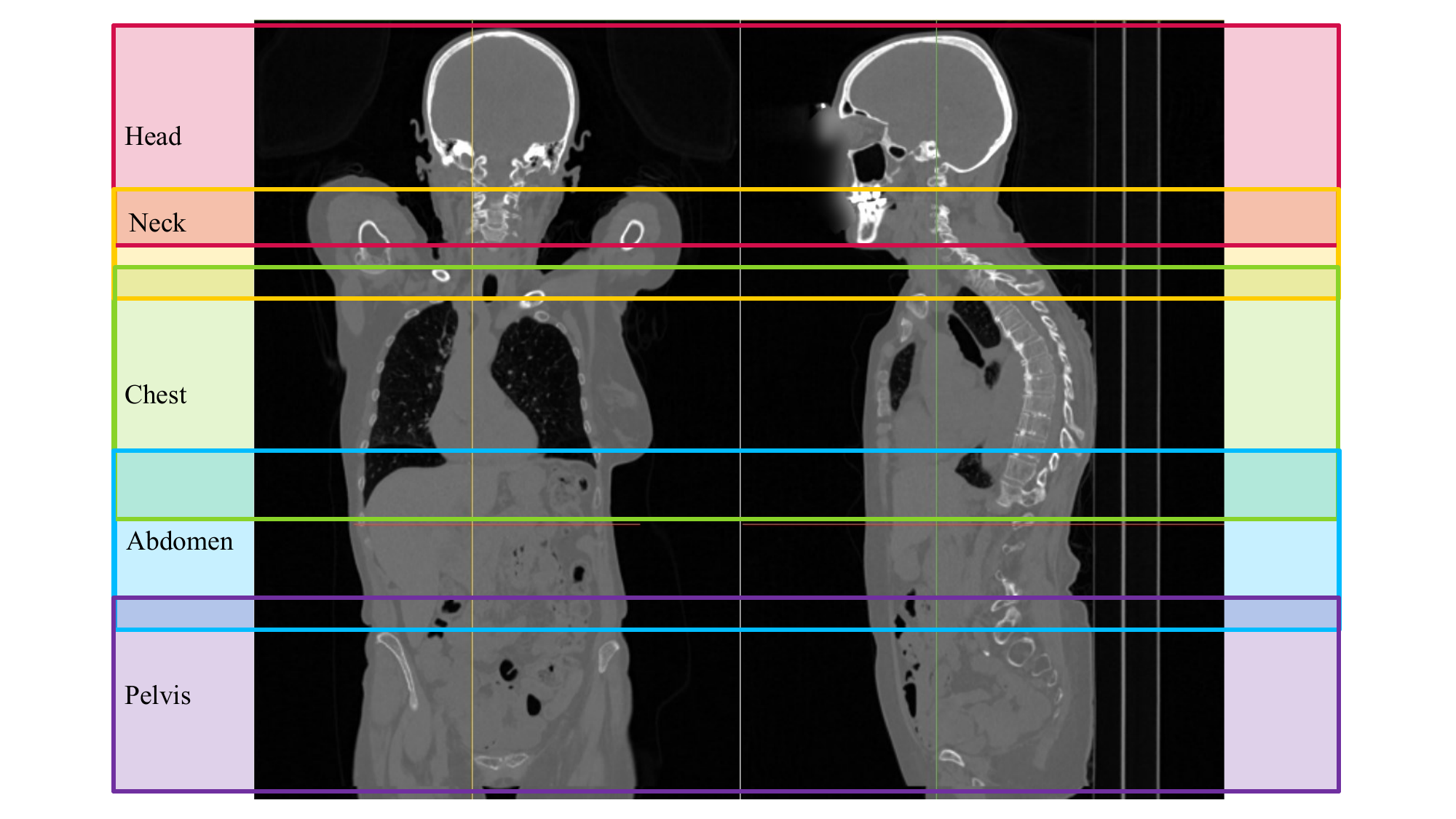}
    \caption{Example anatomical regions delineated on a whole body CT Image (sagittal and coronal view).}
    \label{fig: Mr regions}
\end{figure}

For our rule-based zero-shot methods, precise and radiologist-verified anatomical region definitions are fundamental. As direct body-region labels are typically unavailable in medical imaging, regions are inferred indirectly by mapping the presence of specific organs and identified vertebral levels to predefined anatomical boundaries. These definitions are presented visually in \Cref{fig: CT regions} and \Cref{fig: Mr regions}, and formally detailed in \Cref{tab:regin definition}.
Our system is designed to identify all regions present within a scan, whether single, multiple, or consecutive (e.g., abdomen, chest or chest–abdomen–pelvis, etc.). Importantly, the defined regions are not mutually exclusive; adjacent regions often contain overlapping organs (e.g., liver appearing in both chest and abdomen boundaries). The rule-based system inherently accommodates this by inferring multiple regions simultaneously when their associated organs are detected and sufficiently covered. This modular framework allows for easy adaptation or expansion of regions without model retraining, providing a flexible, transparent, and radiologist-grounded basis for automated body region detection in CT and MRI.

\begin{table}[!hbtp]

\centering
\caption{Radiologist-defined anatomical boundaries.}

\resizebox{\columnwidth}{!}{
\begin{tabular}{lll}
\hline
\textbf{Region} & \textbf{Start} & \textbf{End} \\
\hline
Head & Above the skull & Below the chin / C3--C4 \\
Neck & Skull   & Manubrium sterni / T1--T2 \\
Chest & First rib / C6--C7 & Below the diaphragm / L1 \\
Abdomen & Above diaphragm / T8--T9 & Iliac crests upper edge / L3--L4 \\
Pelvis & Iliac crests upper edge / L3--L4 & Below symphysis pubis \\
\hline
\end{tabular}
}
\label{tab:regin definition}

\end{table}

\begin{figure}[t]
    \centering
    \includegraphics[trim=0.5cm 4.6cm 11.0cm 1cm, clip, width=\columnwidth]{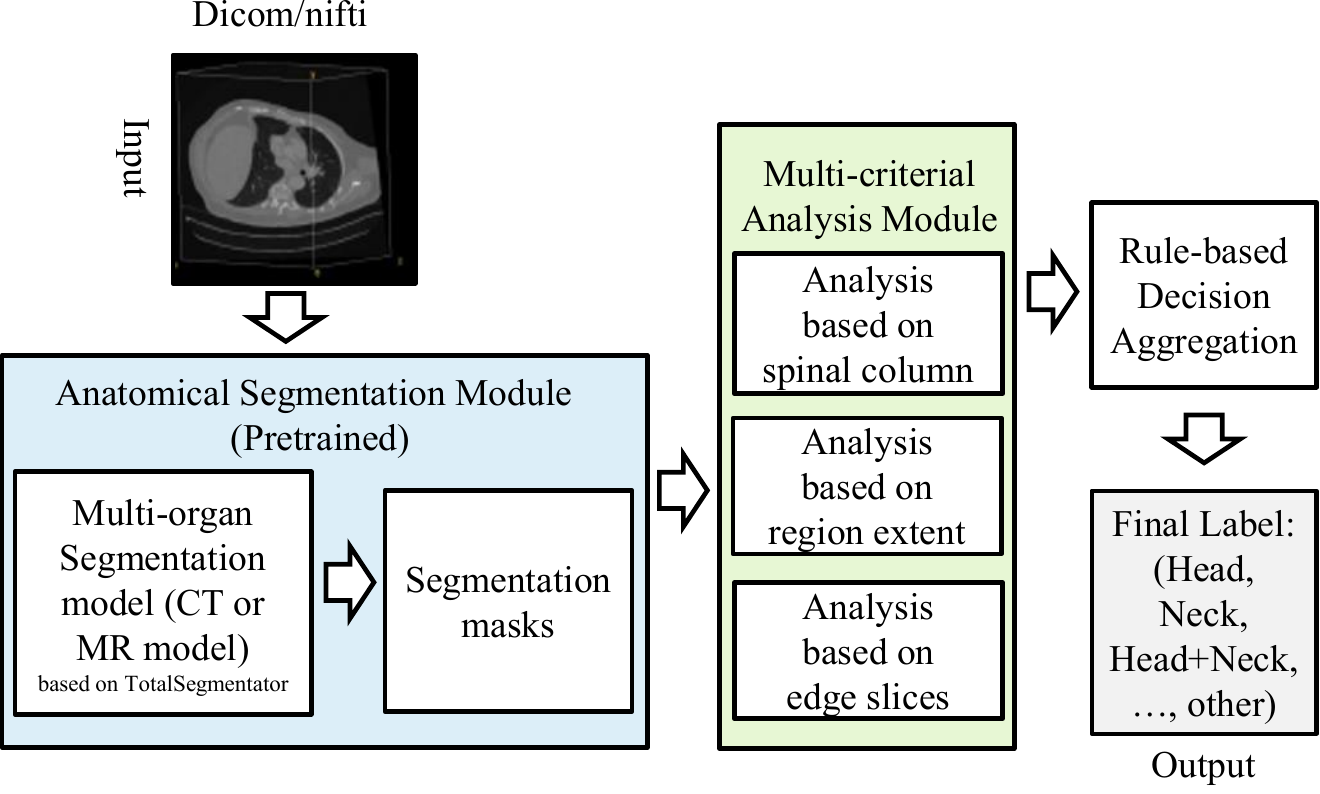}
    \caption{Overview of Zero-shot Segmentation–driven Rule-based approach}
    \label{fig: system1}
\end{figure}

\subsection{Zero-shot Segmentation–driven Rule-based Approach}

\Cref{fig: system1} illustrates our zero-shot segmentation–driven rule-based approach, comprising an anatomical segmentation module, a multi-criterial analysis module, and a rule-based decision aggregation module. The process begins by feeding the input image into a multi-organ segmentation model. We utilize three pre-trained TotalSegmentator models: "total" for CT \cite{wasserthal2023totalsegmentator}, and "total\_mr" and "vertebrae\_mr" for MR \cite{d2024totalsegmentator}. These models provide full-volume multi-organ segmentations (117 classes for CT, 75 for MR; see model repository\footnote{https://github.com/wasserth/TotalSegmentator\#class-details} for full lists), forming the anatomical foundation for all subsequent rule-based decisions. For region detection, only a subset of these classes, selected based on anatomical relevance and radiologist-defined rules, is employed; these are detailed in \Cref{tab:combined_structures} along with their corresponding mapped regions.

Following the segmentation step, a multi-criterial analysis module integrates information from three complementary analytical streams to provide a robust assessment of anatomical coverage. The system first identifies all segmented structures, maps them to their corresponding body regions based on predefined anatomical associations, and then calculates a comprehensive score for each potential region (here, head, neck, chest, abdomen, pelvis).

\begin{itemize}
    \item Spinal Column Analysis (Vertebrae Presence): This criterion provides a hierarchical indicator of vertical anatomical coverage by evaluating the completeness and distribution of segmented vertebral bodies, from C1 through L5 and the sacrum (refer to \Cref{tab:combined_structures}, Vertebrae for details). The system checks for the presence of specific vertebrae defining a region; for instance, C1-C7 for the neck, or L1-L4 for the abdomen, etc. This analysis offers a foundational assessment of spinal coverage, leveraging spinal levels as critical anatomical landmarks.
    
    \item Region Extent Analysis (Organ Extent): This criterion quantifies the cranio-caudal length of each body region based on the presence of its constituent organs. The vertical span defined by the segmented organs is measured and compared against predefined anatomical thresholds. This comparison determines whether the scan adequately covers the expected vertical span of a region, flagging instances of partial or unusually extended anatomical coverage. The constituent organ per modality is defined in \Cref{tab:combined_structures}, Organ/Structures column. 
    
    \item Edge Slice Analysis (Boundary Content): This criterion examines the anatomical content specifically at the superior and inferior boundaries of the scan volume. It identifies organs present at the first and last axial slices containing non-empty anatomical information. This analysis provides crucial information about the overall cranio-caudal range of the scan, which is essential for inferring potential truncation or confirming complete anatomical coverage.
\end{itemize}

\begin{table*}[!t]
    \centering
    \caption{CT and MR class Groups by Region based on TotalSegmentator "total",  "total\_mr" and "vertebrae\_mr" models}

    \begin{tabular}{@{}p{1cm} p{2cm} p{10cm}@{}}
        \toprule
        \textbf{Modality} & \textbf{Region} & \textbf{Organs/Structures} \\ \midrule
        \multirow{5}{*}{CT}   & Head   & brain, skull \\ 
                              & Neck   & esophagus, trachea, thyroid\_gland \\ 
                              & Chest  & lung\_upper\_lobe\_left, lung\_lower\_lobe\_left, lung\_upper\_lobe\_right, lung\_middle\_lobe\_right, lung\_lower\_lobe\_right, heart, rib\_left\_1:12, rib\_right\_1:12, sternum, costal\_cartilages \\ 
                              & Abdomen & spleen, kidney\_right, kidney\_left, gallbladder, liver, stomach, pancreas, duodenum, colon \\ 
                              & Pelvis  & urinary\_bladder, prostate, hip\_left, hip\_right, sacrum \\ 
        \midrule
        \multirow{5}{*}{MR}   & Head   & brain \\ 
                              & Neck   & esophagus \\ 
                              & Chest  & lung\_left, lung\_right, heart \\ 
                              & Abdomen & spleen, kidney\_right, kidney\_left, gallbladder, liver, stomach, pancreas, duodenum, colon \\ 
                              & Pelvis  & urinary\_bladder, prostate, hip\_left, hip\_right, sacrum \\ 
    \end{tabular}
    
    \hspace{1cm} 
    
    \begin{tabular}{@{}p{1cm} p{2cm} p{10cm}@{}}
        \toprule
        \textbf{} & \textbf{Region} & \textbf{Vertebrae} \\ \midrule
        \multirow{5}{*}{CT}     & Head     & vertebrae\_C4:C1 \\ 
                              & Neck     & vertebrae\_T2, vertebrae\_T1, vertebrae\_C7:C1 \\
                              & Chest    & vertebrae\_L1, vertebrae\_T12:T1, vertebrae\_C7, vertebrae\_C6 \\ 
                              & Abdomen  & vertebrae\_L4:L1, vertebrae\_T12:T8 \\ 
                              & Pelvis   & sacrum, vertebrae\_S1, vertebrae\_L5, vertebrae\_L4, vertebrae\_L3 \\ 
        \midrule
        \multirow{5}{*}{MR}     & Head     & vertebrae\_C4:C1 \\ 
                              & Neck     & vertebrae\_T2, vertebrae\_T1, vertebrae\_C7:C1 \\
                              & Chest    & vertebrae\_L1, vertebrae\_T12:T1, vertebrae\_C7, vertebrae\_C6 \\ 
                              & Abdomen  & vertebrae\_L4:L1, vertebrae\_T12:T8 \\ 
                              & Pelvis   & sacrum, vertebrae\_L5, vertebrae\_L4, vertebrae\_L3 \\ 
        \bottomrule
    \end{tabular}
    \label{tab:combined_structures}
\end{table*}

\begin{table*}[!h]
\caption{Quantitative Criteria and Decision Logic for Each Body Region by Modality.}
\label{tab:combined_criteria_final_elegant_fit}
\centering
\resizebox{\textwidth}{!}{
\scriptsize
\begin{tabular}{lllllll}
\hline
\textbf{Body Region} & 
\multicolumn{2}{c}{\textbf{Expected Vertebrae Count (V)}} & 
\multicolumn{2}{c}{\textbf{Expected Organ Count (O)}} & 
\textbf{Region Extent (E)} & 
\textbf{Condition Winning Logic} \\
\cline{2-5}
& \textbf{CT} & \textbf{MR} & \textbf{CT} & \textbf{MR} & \textbf{Range (cm)} & \textbf{(Threshold for 'Win')} \\
\hline
\textbf{Head} & 4 (C1--C4) & 4 (C1--C4) & 2 & 1 & (15, 25) & V $\ge 60\%$; E $\in [70\%, 130\%]$ \\
\textbf{Neck} & 9 (T2, T1, C1--C7) & 9 (T2, T1, C1--C7) & 3 & 1 & (8, 15) & V $\ge 60\%$; min. 1 for CT; E $\in [70\%, 130\%]$ \\
\textbf{Chest} & 15 (L1, T1--T12, C7, C6) & 15 (L1, T1--T12, C7, C6) & 34 & 3 & (20, 35) & V $\ge 60\%$; CT: O $\ge 30\%$, MR: min 1; E $\in [70\%, 130\%]$ \\
\textbf{Abdomen} & 8 (L1--L4, T8--T12) & 8 (L1--L4, T8--T12) & 9 & 9 & (15, 25) & V $\ge 60\%$; O $\ge 30\%$; E $\in [70\%, 130\%]$ \\
\textbf{Pelvis} & 5 (Sac., S1, L3--L5) & 4 (Sac., L3--L5) & 5 & 5 & (15, 25) & V $\ge 60\%$; O $\ge 30\%$; E $\in [70\%, 130\%]$ \\
\hline
\end{tabular}
}
\end{table*}

The final anatomical classification is derived from a Rule-based Decision Aggregation module that quantifies coverage against the parameters in \Cref{tab:combined_criteria_final_elegant_fit}. During condition scoring, each region (e.g., head, neck, etc.) wins a condition if its measured anatomical content meets explicit, modality-specific thresholds. The conditions vary based on the available landmarks for each modality, and it is detailed in \Cref{tab:combined_criteria_final_elegant_fit}, for example, for the abdomen, vertebrae (V) require $\mathbf{\ge 60\%}$ of the expected vertebral classes to be present. Organ Count (O) requires $\mathbf{\ge 30\%}$ of the expected key organs to be segmented, which varies significantly between CT and MR scans. Finally, Region Extent (E) is won if the measured cranio-caudal span falls within $\mathbf{70\%}$ of the nominal minimum extent and $\mathbf{130\%}$ of the nominal maximum extent, ensuring the classification excludes only cases of extreme truncation or overscanning. Thresholds were selected in consultation with a radiologist to tolerate partial coverage while excluding extreme truncation, reflecting common clinical scan variability, and can be adjusted based on the study requirements. This quantitative scoring yields region-specific completeness scores, which the hierarchical labeling module leverages to synthesize the final, anatomically ordered label (e.g., "chest+abdomen"). The final label contains the regions with the highest complete scores (most wins). All the thresholds can be adopted based on the requirements of specific studies or clinical protocols. 

\begin{figure}[t]
    \centering
    \includegraphics[trim=3cm 6.1cm 10cm 0.3cm, clip, width=\columnwidth]{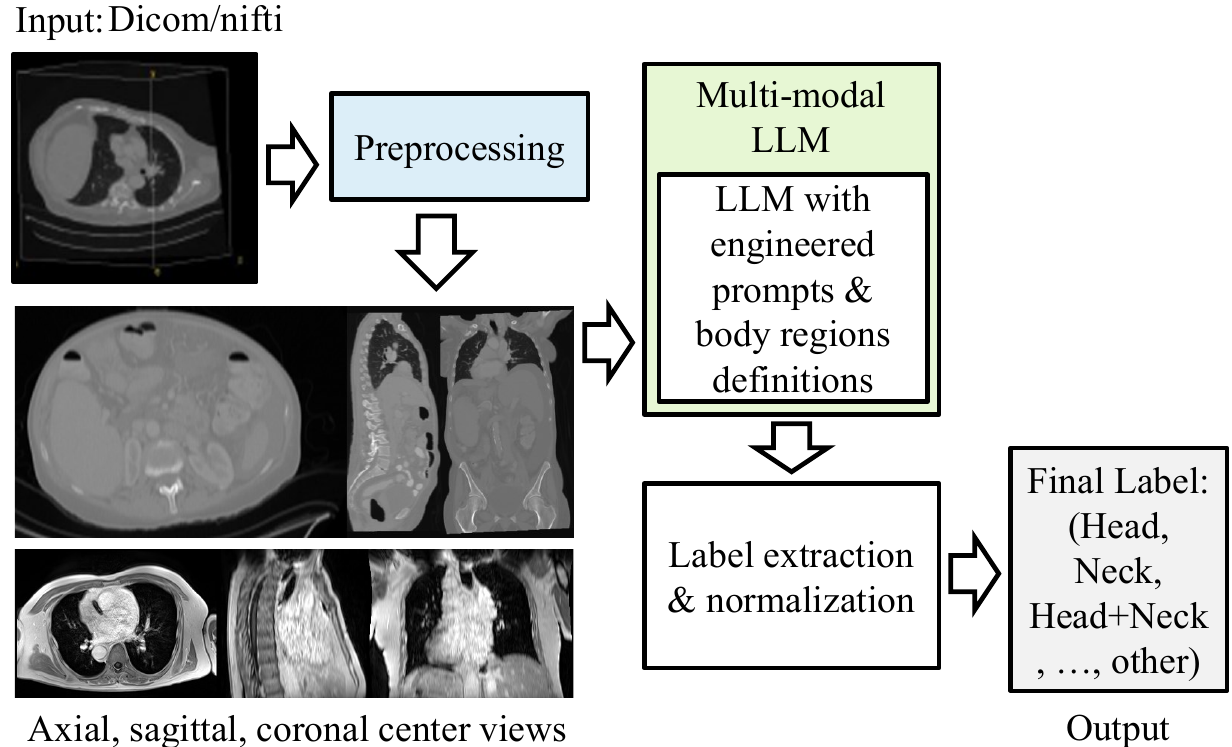}
    \caption{Overview of Zero-shot MLLM approach }
    \label{fig: system2}
\end{figure}

\subsection{Zero-shot MLLM Approach}

The MLLM approach serves as an alternative approach to the segmentation-based method, providing a powerful, rule-guided architecture for anatomical coverage assessment. This pipeline leverages the multimodal capabilities of general LLMs, e.g., GPT or Gemini, to emulate the clinical reasoning of a radiologist viewing a scan across multiple planes (in this study, Gemini 2.5, Flash is used \cite{comanici2025gemini}, but any MLLM can be used for further extensions). 
\Cref{fig: system2} shows an overview. The process begins with input preparation, where the DICOM/NIFTI volume is pre-processed to generate three orthogonal views: Axial, Sagittal, and Coronal. 
These three images are concatenated into a single, composite image, which is then provided as the visual input to the MLLM. The MLLM is directed by a comprehensive system prompt that defines its role as an expert assessor and explicitly injects the structured rules for classification based on organ indication and expected coverages defined in \Cref{tab:combined_structures} and \Cref{tab:combined_criteria_final_elegant_fit}.

The core of the system is the MLLM execution and rule injection, where the model performs cross-view validation to synthesize a single, comprehensive anatomical label. 
The decision-making process is strictly governed by three key criteria, conceptually aligning with the logic used in the segmentation model despite the difference in data input. 
The model is explicitly guided by the same radiologist-defined anatomical boundaries defined in \Cref{tab:regin definition}, such as the chest region spanning from the first rib/C6-C7 down to below the diaphragm/L1, etc. The MLLM is then specifically tasked with applying the defined rules: it is instructed to identify and include only those regions where at least, e.g., 60 percent of the anatomically defined region is visible across the three composite views. This instruction forces the MLLM to conduct a visual and comparative judgment based on the provided cross-sectional images ($\text{Axial}$, $\text{Sagittal}$, and $\text{Coronal}$), effectively serving as a qualitative proxy for the segmentation model's quantitative thresholding. This rule-guided process ensures that the final output, an anatomically ordered label (e.g., $\text{Chest+Abdomen}$), maintains consistency with the criteria for sufficient anatomical coverage.

The success of this rule-guided inference is finalized in the label extraction and normalization module, which is crucial for ensuring consistency and compliance with the overall system requirements. A significant challenge in designing an MLLM workflow is to successfully prompt the model to generate the required structured output, and then parse and extract this information reliably. Once extracted, the label undergoes strict normalization and is validated against a predefined list of allowed anatomical regions (head, neck, chest, abdomen, pelvis, and other). 
This ensures that any model ambiguity or deviation (e.g., classifying a scan as "lower abdomen") is consistently mapped to the appropriate standard label (i.e., abdomen) and formatted in the required anatomical sequence (e.g., abdomen+pelvis) before being saved as the final output. 
Crucially, this MLLM-based system offers significant flexibility: if the classification requirements change, such as the need for new compound labels or the modification of inclusion criteria, only the LLM system prompt requires modification, making the system highly adaptable without the need for model retraining or extensive code changes to the underlying logic.

\begin{figure}[h]
    \centering
    \includegraphics[trim=3.2cm 2cm 7.1cm 2.2cm, clip, width=\columnwidth]{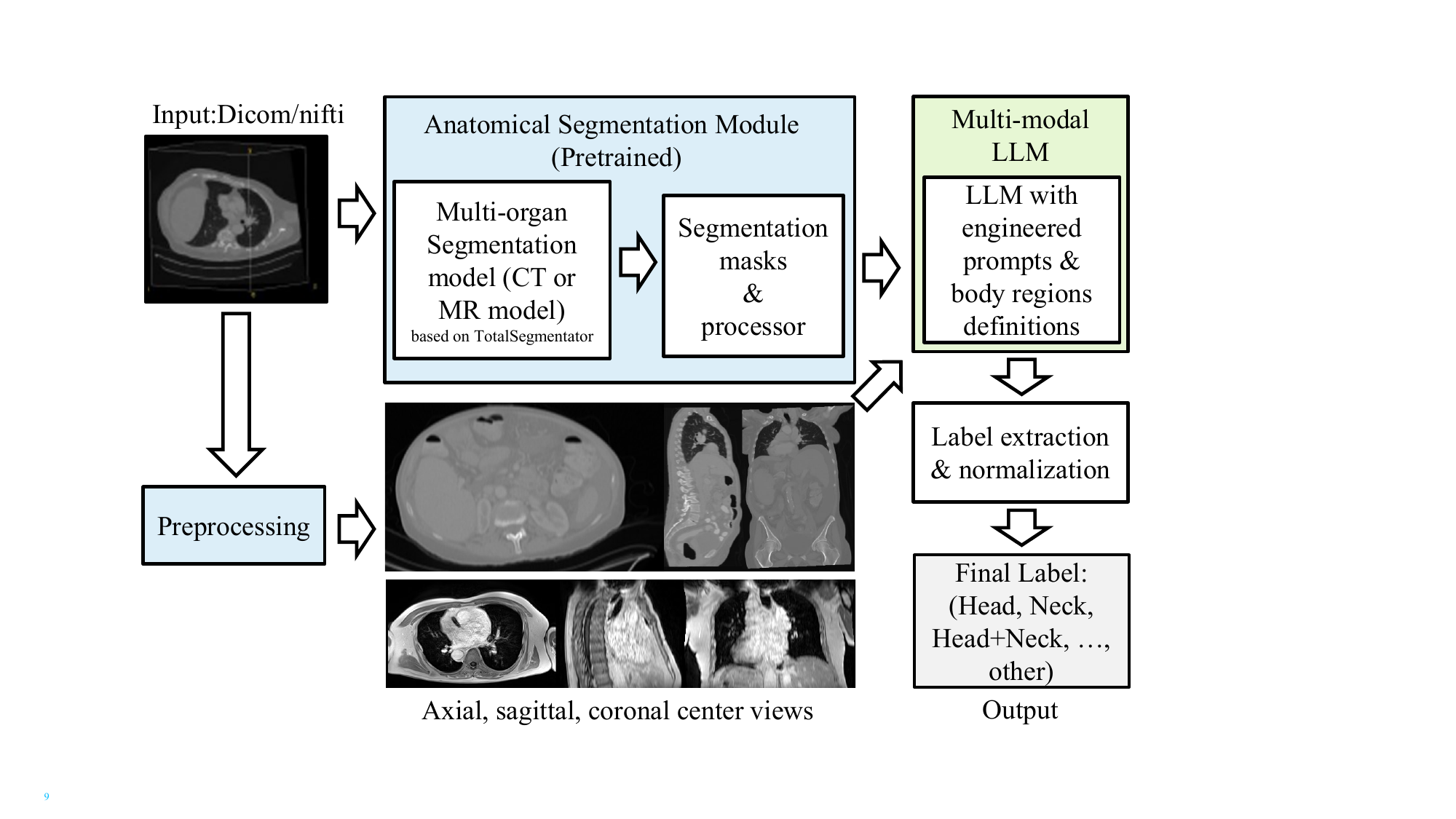}
    \caption{Overview of Segmentation-aware MLLM approach}
\end{figure}

\subsection{Segmentation-aware MLLM Approach}

We introduce a third, hybrid approach: the segmentation-aware MLLM, which aims to transform the MLLM into a segmentation-aware model by supplying it with structured, auxiliary information derived from the initial segmentation pipeline. Along with the composite, cross-sectional views (Axial, Sagittal, Coronal), the model receives the list of segmented organs and their calculated region sizes.
By receiving this explicit anatomical evidence, the MLLM is equipped to apply the instructed logics with greater confidence and precision. 
For instance, when evaluating a potential abdomen region, the MLLM not only visually identifies the liver but also receives the segmented output confirming the presence and the size of the $\text{T8-T9}$ to $\text{L3-L4}$ vertebral levels (from \Cref{tab:regin definition}). This dual input structure (visual cues combined with explicit anatomical facts) is aimed at rendering the MLLM's reasoning more constrained, interpretable, particularly for subtle or ambiguous boundary slices where direct visual judgment is difficult. Crucially, this hybrid approach remains entirely training-free, requiring only modifications to the input prompt format to inject the structured auxiliary data.

\section{Evaluation}

All three methods were evaluated in a completely training-free setting, meaning neither the rule-based system nor the MLLM was trained on body region labels. For comparison, \cite{golzan2025automatic} model is trained in a 5-fold cross-validation (CV) on our labeled dataset. Note that the method proposed by \cite{golzan2025automatic} is specifically designed for CT images; we adopted this method to use average intensity projection for MR images instead of maximum intensity projection.

Our analysis was conducted on 887 medical volumes (465 CT, 422 MR) from the TotalSegmentator public datasets \cite{wasserthal2023totalsegmentator, d2024totalsegmentator}. 
Each scan was manually labeled by a reviewer according to the radiologist-defined anatomical boundaries (\Cref{tab:regin definition}). 
The labels consist of single or combined regions, e.g., abdomen, chest+abdomen, etc.
Scans with atypical coverage that did not satisfy the quantitative criteria of \Cref{tab:combined_criteria_final_elegant_fit} for any primary region (e.g., extremities, heart images) or scans showing only a sliver of an organ or a region were assigned to the "other" category. 
This category serves as a crucial test of each method's specificity and its ability to reject cases that do not conform to the predefined anatomical rules.

Performance was quantified using Accuracy, Precision, Recall, and F1 Score, calculated on a per-region basis. This allowed us to measure how effectively each zero-shot approach could infer the presence of a specific body region. 
The performance of the three zero-shot pipelines on CT and MR scans is detailed in \Cref{tab:per-class-CT} and \Cref{tab:per-class-MR}, respectively.

\begin{table}[!b]
 \caption{Performance metrics for different body region detection methods on CT images. The table compares zero-shot segmentation-driven approach (Segmentation-based), zero-shot MLLM approach, and segmentation-aware MLLM approach (Seg-aware MLLM), and a trained model based on \cite{golzan2025automatic} across various anatomical regions based on Accuracy, Precision, Recall, and F1 Score.}
  \centering
  \resizebox{\columnwidth}{!}{
    \begin{tabular}{cllllll}
    \multicolumn{1}{c}{Region} & Method & \multicolumn{1}{l}{Accuracy} & \multicolumn{1}{l}{Precision} & \multicolumn{1}{l}{Recall} & \multicolumn{1}{l}{F1 Score} & \multicolumn{1}{l}{support}\\
    \hline
   \multirow{4}[0]{*}{Head} & Segmentation-based & 0.973 & 0.879 & 0.967 & 0.921 &  \multirow{4}[0]{*}{60} \\
          & MLLM   & 0.761 & 0.969 & 0.525 & 0.681  &  \\
          & Seg-aware MLLM   & 0.795 & 0.900   & 0.600   & 0.720 & \\
          & Golzan et al. \cite{golzan2025automatic} (5-fold CV)  & 0.947±0.016 & 0.780±0.058  & 0.803±0.123   & 0.785±0.065 &  \\
          \hline
     \multirow{4}[0]{*}{Neck} & Segmentation-based & 0.971 & 0.925 & 0.954 & 0.939  & \multirow{4}[0]{*}{65} \\
          & MLLM   & 0.962 & 0.922 & 0.937 & 0.929 & \\
          & Seg-aware MLLM   & 0.949 & 0.692 & 0.969 & 0.808 & \\
          & Golzan et al. \cite{golzan2025automatic} (5-fold CV)  & 0.967±0.021 &   0.915±0.087  & 0.843±0.107   & 0.871±0.062   &  \\
          \hline
    \multirow{4}[0]{*}{Chest} & Segmentation-based & 0.984 & 1.000     & 0.968 & 0.984 & \multirow{4}[0]{*}{185} \\
          & MLLM   & 0.927 & 0.842 & 0.978 & 0.905 &  \\
          & Seg-aware MLLM   & 0.795 & 0.619 & 1.000 & 0.764 & \\
          & Golzan et al. \cite{golzan2025automatic} (5-fold CV)  & 0.965±0.010  &  0.970±0.040  &  0.936±0.037  & 0.951±0.017   &  \\
          \hline
    \multirow{4}[0]{*}{Abdomen} & Segmentation-based & 0.971 & 0.978 & 0.975 & 0.976 & \multirow{4}[0]{*}{276}\\
          & MLLM   & 0.886 & 0.924 & 0.880  & 0.902 & \\
          & Seg-aware MLLM   & 0.840  & 0.828 & 0.978 & 0.897 & \\
          & Golzan et al. \cite{golzan2025automatic} (5-fold CV)  &  0.927±0.033 &  0.939±0.039   &   0.926±0.049  &  0.931±0.035  &  \\
          \hline
    \multirow{4}[0]{*}{Pelvis} & Segmentation-based & 0.996 & 1.000 & 0.991 & 0.996  & \multirow{4}[0]{*}{230} \\
          & MLLM   & 0.944 & 0.921 & 0.969 & 0.944 & \\
          & Seg-aware MLLM   & 0.979 & 0.958 & 1.000   & 0.979 & \\
          & Golzan et al. \cite{golzan2025automatic} (5-fold CV)  &  0.984±0.005 &  0.992±0.015  &  0.972±0.017   &  0.982±0.006 &  \\
          \hline
    \multirow{4}[0]{*}{Other} & Segmentation-based & 0.961 & 0.800   & 0.941 & 0.865 & \multirow{4}[0]{*}{34}  \\
          & MLLM   & 0.528 & 0.500   & 0.061 & 0.108 & \\
          & Seg-aware MLLM   & - & 0  & 0    & -  & \\
          & Golzan et al. \cite{golzan2025automatic} (5-fold CV)  & 0.958±0.017 &  0.674±0.117  &   0.723±0.158  &  0.6863±0.100  &  \\
    \end{tabular}%
  
}

\label{tab:per-class-CT}%
\end{table}%

\begin{table}[!htbp]
 \caption{Performance metrics for different body region detection methods on MR images. The table compares zero-shot segmentation-driven approach (Segmentation-based), zero-shot MLLM approach, and segmentation-aware MLLM approach (Seg-aware MLLM), and a trained model based on \cite{golzan2025automatic} across various anatomical regions based on Accuracy, Precision, Recall, and F1 Score.}
  \centering
  \resizebox{\columnwidth}{!}{
    \begin{tabular}{cllllll}
    \multicolumn{1}{c}{Region} & Method & \multicolumn{1}{l}{Accuracy} & \multicolumn{1}{l}{Precision} & \multicolumn{1}{l}{Recall} & \multicolumn{1}{l}{F1 Score} & \multicolumn{1}{l}{Support}  \\
    \hline
    \multirow{4}[0]{*}{Head} & Segmentation-based & 0.940 & 0.978 & 0.882 & 0.928 & \multirow{4}[0]{*}{51} \\
          & MLLM   & 0.957 & 0.940 & 0.922 & 0.931 & \\
          & Seg-aware MLLM   & 0.959 & 0.855 & 0.940 & 0.895 &  \\
          & Golzan et al. \cite{golzan2025automatic} (5-fold CV)  & 0.974±0.021 &  0.928±0.092  & 0.800±0.134  &   0.855±0.101 &  \\
          \hline
    \multirow{4}[0]{*}{Neck} & Segmentation-based & 0.857 & 0.900 & 0.720 & 0.800  & \multirow{4}[0]{*}{25} \\
          & MLLM   & 0.751 & 0.304 & 0.583 & 0.400 & \\
          & Seg-aware MLLM   & 0.790 & 0.320 & 0.667 & 0.432 &  \\
          & Golzan et al. \cite{golzan2025automatic} (5-fold CV)  & 0.966±0.017  &  0.800±0.245 &  0.563±0.127 & 0.6276±0.101 &  \\
          \hline
    \multirow{4}[0]{*}{Chest} & Segmentation-based & 0.993 & 0.921 & 1.000 & 0.959 & \multirow{4}[0]{*}{58} \\
          & MLLM   & 0.927 & 0.776 & 0.897 & 0.832 &  \\
          & Seg-aware MLLM   & 0.914 & 0.483 & 1.000 & 0.652 &  \\
         & Golzan et al. \cite{golzan2025automatic} (5-fold CV)  & 0.988±0.007  &  0.984±0.031 & 0.907±0.095 &  0.940±0.048 &  \\
          \hline
    \multirow{4}[0]{*}{Abdomen} & Segmentation-based & 0.972 & 0.979 & 0.949 & 0.964 & \multirow{4}[0]{*}{99}  \\
          & MLLM   & 0.924 & 0.897 & 0.879 & 0.888 & \\
          & Seg-aware MLLM   & 0.891 & 0.613 & 0.969 & 0.751 &  \\
         & Golzan et al. \cite{golzan2025automatic} (5-fold CV)  & 0.966±0.016  &  0.912±0.090 & 0.912±0.060 & 0.908±0.042  &  \\
          \hline
    \multirow{4}[0]{*}{Pelvis} & Segmentation-based & 0.933 & 0.898 & 0.883 & 0.891 & \multirow{4}[0]{*}{60} \\
          & MLLM   & 0.901 & 0.607 & 0.900 & 0.725 & \\
          & Seg-aware MLLM   & 0.908 & 0.604 & 0.917 & 0.728 & \\
          & Golzan et al. \cite{golzan2025automatic} (5-fold CV)  & 0.956±0.014 & 0.867±0.121 & 0.758±0.120 & 0.797±0.073 &  \\
          \hline
    \multirow{4}[0]{*}{Other} & Segmentation-based & 0.946 & 0.939 & 0.952 & 0.945 & \multirow{4}[0]{*}{209}  \\
          & MLLM   & 0.895 & 0.988 & 0.800 & 0.884 &  \\
          & Seg-aware MLLM   & 0.852 & 0.993 & 0.709 & 0.827 & \\
          & Golzan et al. \cite{golzan2025automatic} (5-fold CV)  & 0.916±0.026 & 0.902±0.041 & 0.899±0.055  & 0.899±0.033&  \\
    \end{tabular}%
  
  }
    \label{tab:per-class-MR}%
\end{table}%

\subsection{Zero-shot Segmentation–driven Rule-based Approach} 
This approach establishes a high-performance baseline. 
The zero-shot segmentation-driven rule-based approach was the most consistently accurate and reliable method across both modalities. 
For CT scans (\Cref{tab:per-class-CT}), it delivered the best F1 scores for all major anatomical regions, including chest ($0.984$), abdomen ($0.976$), and pelvis ($0.996$). 
Its performance on MR scans (\Cref{tab:per-class-MR}) was similarly strong, achieving top scores in most categories, including an F1 of $0.964$ for abdomen, $0.959$ for chest, and $0.928$ for head. 
Furthermore, this method proved robust in handling scans with atypical coverage or undefined regions, correctly identifying "other" cases with high F1 scores of $0.865$  for CT and $0.945$ for MR. This demonstrates the effectiveness of its quantitative, multi-criterial logic.
In our experiment setup, this zero-shot approach outperforms the trained state-of-the-art model based on \cite{golzan2025automatic} without any requirement for training data.

\subsection{Zero-shot MMLLM-based Approach} 

The generic MLLM, which relies only on visual interpretation guided by a textual prompt, demonstrated its capability as a powerful, training-free alternative. 
It delivered highly competitive performance across several key anatomical regions, particularly where strong visual patterns were present. In CT scans (\Cref{tab:per-class-CT}), it achieved F1 scores of $0.929$ for neck, $0.905$ for chest, $0.902$ for abdomen, and $0.944$ for pelvis, often approaching the performance of the specialized rule-based system. Its aptitude for visual reasoning was most evident in MR scans (\Cref{tab:per-class-MR}), where it achieved the top-performing F1 score for the head region ($0.931$) and handled atypical "other" cases effectively with a strong F1 score of $0.884$.

However, the model's reliance on generalized visual understanding also led to specific, notable weaknesses. In contrast to its success on the MR head, its performance was significantly weaker on the CT head, where the F1 score was only $0.681$. 
The most significant limitation was its lack of specificity in certain contexts; the model struggled to reject scans that did not fit the predefined criteria in CT, resulting in a near-complete failure on the CT "other" category with an F1 score of just $0.108$. Lastly, while its performance on MR neck was low ($0.400$ F1), this was a challenging region for all evaluated methods, indicating a broader difficulty rather than a unique failure of the MLLM.

\subsection{Segmentation-aware MMLLM-based Approach}

The segmentation-aware MLLM was designed to test the hypothesis that providing explicit anatomical evidence would ground the MLLM's visual reasoning and improve its accuracy. However, our results show the opposite was often true: rather than helping, the additional information appeared to confuse the model, revealing a significant weakness in its zero-shot reasoning capabilities. The dominant pattern observed was a drastic trade-off where the model achieved perfect or near-perfect recall at the expense of a catastrophic drop in precision.

This failure mode was most evident in the chest region for both CT and MR. In these cases, the recall of the hybrid model increased to $1.00$, but its precision decreased to $0.619$ (CT) and $0.483$ (MR), respectively. 
This indicates that once the model was told that chest-related organs were present, it began to over-predict the "chest" label, losing the nuanced ability to assess visual completeness that the vanilla MLLM had. While there were marginal F1 score improvements in some areas, such as CT head (from $0.681$ to $0.720$), these minor gains were overshadowed by the widespread negative impact on precision. This experiment demonstrates that simply injecting factual text into a prompt does not guarantee better performance; without task-specific fine-tuning, a generic LLM may not know how to properly weigh or integrate multimodal information, leading to predictable and detrimental behaviors like over-prediction.

\section{Discussion}

This study investigated the feasibility of zero-shot body region detection by repurposing knowledge from large pre-trained models. 
The results highlight a clear trade-off between the reliability of explicit rule-based systems and the flexibility of general-purpose MLLMs. 
The performance of the segmentation-driven rule-based approach confirms that for well-defined tasks, a deterministic approach grounded in high-quality anatomical evidence is superior. Its success, particularly in CT, where anatomical contrast is high, underscores that robust organ and vertebral segmentation is a sufficient foundation for accurate region detection. 
This method’s transparency and reliability make it a prime candidate for integration into automated clinical workflows where correctness and efficiency are paramount. 
The modularity of its rules also allows for easy adaptation to new region definitions without any model retraining.

In contrast, the MLLM-based approach demonstrated the promise and pitfalls of generalized visual reasoning. 
Its ability to achieve high accuracy in certain regions (e.g., MR head) without any explicit anatomical data is remarkable and points to a future where such models could handle tasks with subtle visual cues. 
However, its failures, especially the inability to handle "other" cases in CT, reveal a critical limitation: a lack of specificity. 
Without being explicitly trained on what not to find, the generic LLM tends to force a fit to the categories it knows, a significant risk in medical applications.

Our most insightful finding came from the segmentation-aware hybrid approach. Contrary to our initial hypothesis, providing factual evidence did not improve MLLM reasoning.
Instead, the model's performance showed limitations in reliable multimodal evidence integration under zero-shot constraints in sequential thinking and contextual understanding, especially when faced with nuanced details. 
Rather than using the provided organ list as a clue to guide a careful visual verification of anatomical boundaries, the MLLM appeared to take a simplistic shortcut, leading to over-prediction whenever characteristic organs were present in the text prompt, leading to over-prediction and a sharp decline in precision. 
This failure revealed limitations under strict zero-shot constraints that a generic LLM might lack the intrinsic capability for multi-step reasoning; it struggles to properly weigh or fuse different information sources (its own visual analysis versus explicit textual facts). This finding raises an open question for the practical application of foundation models: how to enable them to coherently integrate diverse data inputs and perform robust sequential reasoning on specialized, zero-shot tasks.

\section{Limitation and Future Work}

A key limitation of our work is the direct dependency on the performance of the upstream segmentation models. Any failure in the segmentation of a critical organ or vertebra directly impacts the accuracy of the rule-based and hybrid systems. Furthermore, the performance on the MR neck class was uniformly low across all methods, suggesting that the available segmentation classes in the pre-trained MR model may be insufficient for defining this region robustly. Future work should explore using outputs from multiple, diverse segmentation models to create a more fault-tolerant ensemble.
Additionally, our study found that extensive manual prompt engineering was insufficient to correct the reasoning failures observed in the MLLM, particularly in the hybrid model. While this highlights limitations of current zero-shot capabilities in this setup, it motivates exploring more advanced methods. Therefore, a promising direction for future work is to investigate few-shot fine-tuning or automated prompt optimization techniques to explicitly teach the model how to properly integrate anatomical evidence without requiring a large labeled dataset.

\section{Conclusion}

In this paper, we successfully demonstrated that accurate, metadata-independent body region detection can be achieved in a completely zero-shot manner. We presented and evaluated three distinct training-free pipelines: a segmentation-driven rule-based model, a pure multimodal LLM, and a hybrid combination of the two. Our findings show that the rule-based approach, which translates anatomical segmentations into decisions, provides the most reliable, efficient, and transparent solution for immediate practical use.  General-purpose MLLMs also demonstrated impressive flexibility and achieved high performance in specific contexts, proving to be a reliable alternative for certain use cases. However, their variable specificity suggests that broader medical deployment in a zero-shot setting requires careful task selection or further model fine-tuning. This work not only provides a practical solution for automating a critical step in medical image analysis but also sheds light on the current capabilities and limitations of leveraging large pre-trained models for specialized, zero-shot tasks in medicine.
From a clinical deployment perspective, the segmentation-driven approach offers a favorable balance between accuracy, interpretability, and computational efficiency. Explicit anatomical rules allow for straightforward validation and facilitate regulatory alignment, which remains challenging for generative LLM-based approaches.

\addtolength{\textheight}{-12cm}   





\bibliographystyle{IEEEtran}
\bibliography{ref}

\end{document}